\documentclass[letterpaper]{article} 
\usepackage[table,xcdraw]{xcolor}

\usepackage[draft]{aaai2026}  
\usepackage{times}  
\usepackage{helvet}  
\usepackage{courier}  
\usepackage[hyphens]{url}  
\usepackage{graphicx} 
\usepackage{natbib}  
\usepackage{caption} 
\usepackage{booktabs} 
\usepackage{amsmath} 
\usepackage{multirow}
\usepackage{pifont}
\usepackage{makecell}
\usepackage{amssymb}
\usepackage{algpseudocode}
\usepackage{algorithm}
\usepackage{algorithmicx}

\usepackage{newfloat}
\usepackage{listings}
\DeclareCaptionStyle{ruled}{labelfont=normalfont,labelsep=colon,strut=off} 
\floatstyle{ruled}
\newfloat{listing}{tb}{lst}{}
\floatname{listing}{Listing}
\title{S2-UniSeg: Fast Universal Agglomerative Pooling for \\ Scalable Segment Anything without Supervision}
\author{
    Huihui Xu\textsuperscript{\rm 1, 2},
    Jin Ye\textsuperscript{\rm 1},
    Hongqiu Wang\textsuperscript{\rm 2},
    Changkai Ji\textsuperscript{\rm 1},
    Jiashi Lin\textsuperscript{\rm 1},
    Ming Hu\textsuperscript{\rm 1},\\
    Ziyan Huang\textsuperscript{\rm 1},
    Ying Chen\textsuperscript{\rm 1},
    Chenglong Ma\textsuperscript{\rm 1},
    Tianbin Li\textsuperscript{\rm 1},
    Lihao Liu\textsuperscript{\rm 1},\\
    Junjun He\textsuperscript{\rm 1, 4, $\dagger$},
    Lei Zhu\textsuperscript{\rm 2, 3, $\dagger$}
}
\affiliations{
    \textsuperscript{\rm 1}Shanghai Artificial Intelligence Laboratory\\
    \textsuperscript{\rm 2}The Hong Kong University of Science and Technology (Guangzhou)\\
    \textsuperscript{\rm 3}The Hong Kong University of Science and Technology\\
    \textsuperscript{\rm 4}Shanghai Innovation Institute\\
    hejunjun@pjlab.org.cn, leizhu@ust.hk
}

\usepackage{bibentry}
\begin{document}

\maketitle

\setcounter{footnote}{0}
\renewcommand{\thefootnote}{\fnsymbol{footnote}}
\footnotetext{$^\dagger$ Corresponding authors.}
\begin{abstract}
Recent self-supervised image segmentation models have achieved promising performance on semantic segmentation and class-agnostic instance segmentation.
However, their pretraining schedule is multi-stage, requiring a time-consuming pseudo-masks generation process between each training epoch. 
This time-consuming offline process not only makes it \textbf{difficult to scale with training dataset size}, 
but also leads to \textbf{sub-optimal solutions} due to its discontinuous optimization routine.
To solve these, we first present a novel pseudo-mask algorithm, Fast Universal Agglomerative Pooling (UniAP). 
Each layer of UniAP can identify groups of similar nodes \textit{\textbf{in parallel}}, 
allowing to generate both \textbf{semantic-level and instance-level} and \textbf{multi-granular} pseudo-masks within \textit{\textbf{tens of milliseconds}} for one image. 
Based on the fast UniAP, we propose the Scalable Self-Supervised Universal Segmentation (S2-UniSeg), which employs a student and a momentum teacher for continuous pretraining.
A novel \textbf{segmentation-oriented pretext task}, Query-wise Self-Distillation (QuerySD), is proposed to pretrain S2-UniSeg to learn the local-to-global correspondences.
%
Under the same setting, S2-UniSeg outperforms the SOTA UnSAM model, achieving notable improvements of AP+6.9 on COCO, AR+11.1 on UVO, PixelAcc+4.5 on COCOStuff-27, RQ+8.0 on Cityscapes. After scaling up to a larger 2M-image subset of SA-1B, S2-UniSeg further achieves performance gains on all four benchmarks.
\end{abstract}
\begin{links}
  \link{Code}{https://github.com/bio-mlhui/S2-UniSeg}
\end{links}

\begin{figure}[t]
\begin{center}
\includegraphics[width=\linewidth]{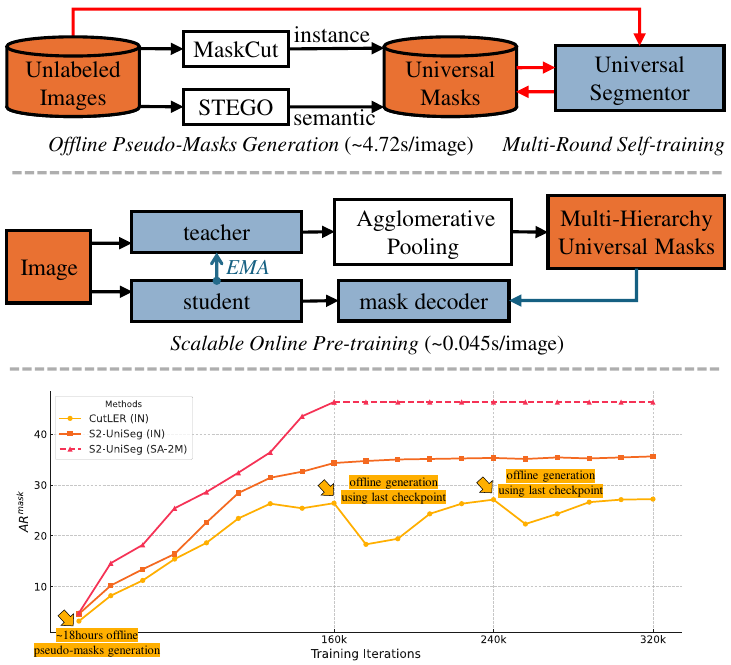}
\end{center}
\caption{\textbf{Previous paradigms vs. our paradigm.} S2-UniSeg eliminates the need for time-consuming offline pseudo-mask generation, functioning as a single-stage pretraining framework with continuous optimization. 
}
\label{fig:comparision}
\end{figure}

\section{Introduction}
Supervised image segmentation has made significant progress in recent years~\cite{segformer, mask2former, sam, seem, maskdino, sam2, xu2024lgrnet, wang2025serp, wang2024language, xu2025medground}.
However, training such segmentation models typically relies on large-scale human annotations~\cite{wu2024rainmamba, tian2023deep, ning2025retinalogos}, which are both time-consuming and labor-intensive. For example, annotating a single image in the SA-1B dataset~\cite{sam} can take over 25 minutes of detailed effort.
Moreover, human annotations are probably noisy, inconsistent, and influenced by subjective biases, leading to misalignment with fine-grained image details and object boundaries. These limitations can undermine the robustness and generalizability of segmentation models across diverse visual contexts.
Therefore, we propose an efficient single-stage self-supervised segmentation model leveraging intrinsic image information to reduce reliance on expensive annotation.

\begin{table}[t]
\begin{center}
\resizebox{\linewidth}{!}{
\begin{tabular}{cccc|cc}
\toprule
Method  & Time (seconds) & $\text{AP}^{\text{mask}}$ & mIoU & Universal & Multi-granular\\
\hline
Divide-and-Conquer~\cite{unsam}  & 5.27 & 7.6 & -         & \ding{56} & \ding{52} \\
TokenCut~\cite{tokencut}         & 2.25 & 3.5 & -         & \ding{56} & \ding{56} \\
MaskCut~\cite{cutler}            & 4.72 & 6.7 & -         & \ding{56} & \ding{56} \\
MaskCut+STEGO~\cite{u2seg}       & 4.86 & 6.7 & 26.3      & \ding{52} & \ding{56} \\
\hline
Universal Agglomerative Pooling  & \textbf{0.045} & 6.2 & 21.8     & \ding{52} & \ding{52} \\
\toprule
\end{tabular}}
\end{center}
\caption{\textbf{Quantitative efficiency and effectiveness comparison among different pseudo-masks generation algorithms.} We run all methods on COCOval2017 with the same hardware settings, and evaluate their average processing time per image, zero-shot class-agnostic instance segmentation performance~($\text{AP}^{\text{mask}}$), unsupervised semantic segmentation performance~(mIoU). All methods use the DINO pretrained ViT-base/8 features. 
}
\label{tab:intro}
\end{table}

As shown in the top of Figure~\ref{fig:comparision}, existing self-supervised image segmentation methods typically adopt a two-stage pipeline~\cite{cutler, unsam, u2seg, cuvler}. In the first stage, based on pretrained self-supervised visual representations~\cite{dino, byol, simsiam, simclr, moco, dinov2}, graph partitioning algorithms~\cite{cutler, tokencut} are used to generate pseudo-masks \textit{offline} for all unlabeled images. These pseudo-masks are then used to initialize training of the segmentation model. In the second stage, segmentation performance is improved via \textit{multi-round self-training}, where pseudo-masks generated from the model checkpoint of the previous round are used to fine-tune the model at next round until convergence.

As shown in Table~\ref{tab:intro}, although the pseudo-masks used in the first stage achieve a higher $\text{AP}^{\text{mask}}$, the generation process is time-consuming. For UnSAM~\cite{unsam}, average time for generating pseudo masks for a 512$\times$512 image takes 5.27 seconds, which is impractical to scale with large-scale datasets. 
Furthermore, as shown in the bottom of Figure~\ref{fig:comparision}, the overall optimization is discontinuous, leading to unstable and sub-optimal convergence.

To achieve scalable and continuous training, the primary challenge is to reduce the time for pseudo-mask generation. To this end, we propose the \textbf{Fast Universal Agglomerative Pooling (UniAP)} to efficiently generate both universal and multi-granular pseudo-masks of a single image within tens of milliseconds. Our key insight is that pixels belonging to the same instance or semantic category are spatially adjacent in form of a connected region. These groups of strongly connected pixels can be identified in parallel~\cite{connected_components}. As shown in Table~\ref{tab:intro}, UniAP achieves comparable segmentation performance while being approximately $100\times$ faster than previous methods, making it feasible for single-stage pretraining and scalable to large-scale dataset.

Equipped with fast pseudo-masks generation by UniAP, we propose the \textbf{Scalable Self-Supervised Universal Segmentation (S2-UniSeg)},
which leverages the teacher-student framework trained with a novel segmentation-oriented pretext task \textbf{Query-wise Self-Distillation (QuerySD)}.
Unlike Self-supervised representation models which utilize global pooling or the ViT~\cite{vit}\texttt{[CLS]} token, we condense each image into a set of universal object queries, and train each student query to predict the corresponding bipartite-matched teacher query. 
Our main contributions are:
\begin{itemize}
    \item We propose a novel and efficient pseudo-masks generation algorithm, Fast Universal Agglomerative Pooling (UniAP), which can generate both universal and multi-granular pseudo-masks within tens of milliseconds. UniAP makes it feasible for continuous optimization and scalable with large-scale dataset.
    \item We propose the Scalable Self-Supervised Universal Segmentation (S2-UniSeg) framework. A novel segmentation-oriented pretext task, Query-wise Self-distillation (QuerySD) is proposed to train a student and a momentum teacher in self-supervised manner. 
    \item Extensive experiments on ImageNet, SA-1B, COCO, UVO, COCOStuff, and Cityscapes demonstrate that S2-UniSeg achieves substantial performance improvements on four universal segmentation tasks. Moreover, S2-UniSeg demonstrates strong scalability with increasing training data, yielding better performance as the dataset size grows.
\end{itemize}

\section{Related Work}
\textbf{Self-supervised Representation Learning (SSL).} Self-supervised representation learning aims to learn general-purpose~\cite{bengio2013representation} features from large amounts of unlabeled data samples without manual annotation. A pretext task is often defined to train the model. According to types of pretext task, they can be classified into \textit{contrastive learning} methods and \textit{masked image modeling} methods. Contrastive learning based methods include pretext tasks based on negative samples~\cite{simclr, moco, mocov3}, clustering~\cite{swav, sela}, self-distillation~\cite{dino, byol, simsiam, dinov2}, and feature decorrelation~\cite{barlow, vicreg}. Masked image modeling methods include pretext tasks based on low-level targets~\cite{mae, simim, maskfeat}, high-level targets~\cite{beit, peco, cae}, self-distillation~\cite{sdae, data2vec}, and multi-modal teacher~\cite{mimco, beitv2}. 

\noindent\textbf{Self-supervised Instance and Semantic Segmentation.} Recently studies~\cite{dino, stego, lost, vo2020toward, vo2021large} show pretrained SSL features can capture pixel-to-pixel semantic similarity. Inspired by that, several works~\cite{stego, cutler, unsam, hp, eagle, tokencut, van2022discovering, wang2022freesolo, alignseg} aim to distill or self-train a segmentation model based on the pretrained SSL representations. These methods can be classified into semantic segmentation methods~\cite{stego, eagle, alignseg,seong2024progressive}, zero-shot class-agnostic instance segmentation methods~\cite{cutler, unsam}, and universal segmentation methods~\cite{u2seg}. State-of-the-art unsupervised zero-shot instance segmentation methods~\cite{cutler, cuvler, unsam} adopt a cut and learn pipeline, in the sense that they first generate pseudo-masks of the whole dataset using pretrained SSL features, and then learn a model through multi-round self-training. Unsupervised semantic segmentation methods~\cite{stego, hp, eagle, alignseg} adopts a distillation-based objective, in the sense that the projected segmentation features should preserve the pixel-to-pixel semantic correspondence in the frozen SSL representation space. Recently, U2Seg~\cite{u2seg} proposes a self-supervised universal segmentation framework for class-aware instance and panoptic segmentation. U2Seg adopts a similar cut-and-learn pretraining pipeline but also clusters the masks to generate their pseudo classification labels.

\noindent\textbf{Graph Pooling.} As an essential component of Graph Neural Networks, graph pooling~\cite{graph_pool, buterez2022graph, grattarola2022understanding, mesquita2020rethinking} is crucial for aggregating information from multiple nodes to obtain holistic subgraph-level representations. Graph pooling can be roughly divided into flat pooling and hierarchical pooling. Flat pooling~\cite{graph_pool1, graph_pool2, graph_pool3}, also known as Graph Readout~\cite{buterez2022graph}, aims to obtain a global representation. Hierarchical pooling aims to iteratively coarsen the graph into smaller size and generate multi-scale graph features. According to how new nodes are formed, it can be classified into node clustering pooling~\cite{graph_pool7, graph_pool8} and node drop pooling~\cite{graph_pool4, graph_pool5, graph_pool6}. The primary distinction lies in that node clustering pooling aggregates information from old nodes into new nodes, whereas node drop pooling directly drops unwanted ones and retains a subset of old nodes.


\section{Scalable Self-Supervised Universal Segmentation}
\begin{figure*}[ht]
\begin{center}
\includegraphics[width=\linewidth]{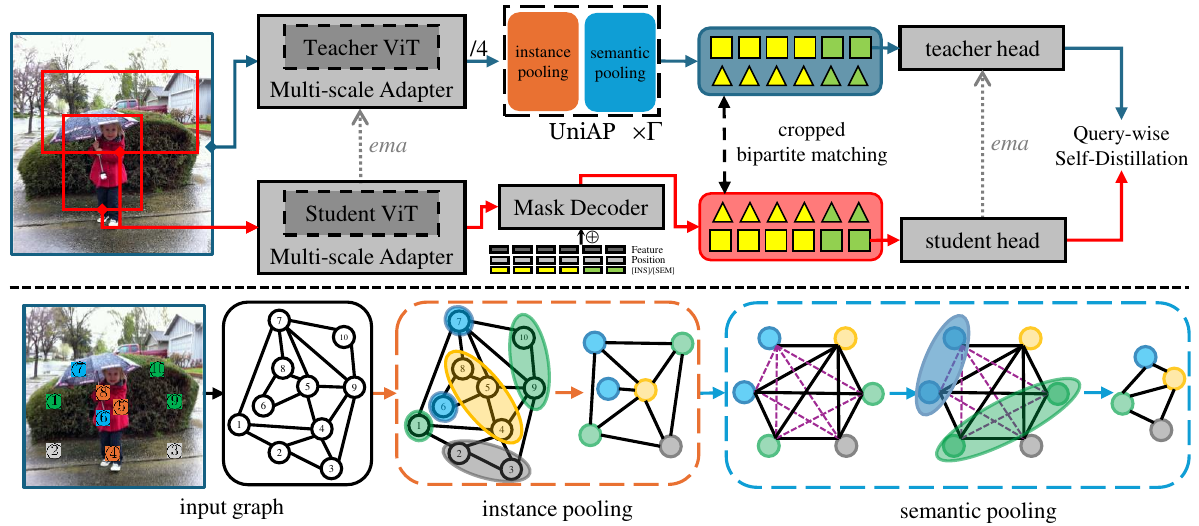}
\end{center}
\vspace{-0.4cm}
\caption{\textbf{Model architecture for Scalable Self-Supervised Universal Segmentation (S2-UniSeg).} (Top) A student and a momentum teacher is leveraged to facilitate self-distillation. The original view (global) is fed into the teacher branch, then we propose Fast Universal Agglomerative Pooling to efficiently generate universal masks ($\triangle$) and their query features ($\square$). In student branch, we use mask decoder with task-specific object queries to predict the local view universal masks. During training, bipartite matching is devised to match student masks with the cropped teacher masks. (Bottom) Each UniAP layer is composed of a semantic and instance pooling obligated to generate universal pseudo-masks. UniAP can efficiently identify groups of strongly-connected nodes \textit{in parallel}, thus enabling single-stage continuous pretraining.}
\label{fig:overview}
\end{figure*}
\subsection{Overview Architecture}
As shown in Figure~\ref{fig:overview}, S2-UniSeg has a student branch and a teacher branch.
Given an image, we follow multi-crop~\cite{swav} to get a set of $\delta$ local views input to the student branch. The original image is treated as the global view input to the teacher branch. 
The student and teacher encoder both adopt the same architecture of a multi-scale encoder~\cite{vit_adapter}. 

\noindent\textbf{Teacher Branch.} For the global view, its feature map with the largest scale (stride 4) is input to a stack of $\Gamma$ UniAP layers to generate the instance and semantic pseudo-masks along with their feature embeddings. We call these embeddings as teacher queries in later sections.

\noindent\textbf{Student Branch.} For each local view, a mask decoder~\cite{mask2former} takes its multi-scale feature maps as input and predicts its semantic and instance masks each along with the query features at the last decoder layer. During query initialization, we partition object queries into two groups—semantic queries, which are distinguished by a learnable token \texttt{[SEM]} and obligated to attend to scattered semantic regions, and instance queries, which are distinguished by a learnable token \texttt{[INS]} and obligated to identify individual instances. Compared with the previous tedious two-branch decoder design in U2Seg~\cite{u2seg}, our framework achieves task distinction using only two learnable tokens. 

\subsection{Fast Universal Agglomerative Pooling}
\label{sec:merge}
As shown in Figure~\ref{fig:overview}, each UniAP layer first identifies groups of strongly-connected nodes, and merges each group into one supernode. Compared with the optimization-based TokenCut~\cite{tokencut} and MaskCut~\cite{cutler}, UniAP is a heuristic nonparametric approach and does not require computation of eigenvectors. Moreover, based on the SCC~\cite{connected_components} algorithm, UniAP can merge a group of strongly-connected nodes \text{in parallel}. As shown in Table~\ref{tab:intro}, UniAP is much faster than existing methods and achieves comparable performance, which can be integrated into each training step to enable single-stage pretraining and continuous optimization routine. 

Moreover, UniAP can also generate pseudo-masks for scattered regions of semantic segmentation, which are not supported in MaskCut~\cite{tokencut}, TokenCut~\cite{cutler}, and Divide-and-Conquer~\cite{unsam}. UniAP is nonparametric and does not require in-domain pretrained STEGO~\cite{stego} as in U2Seg.
Next, we illustrate the graph initialization, identify step, and merge step in detail. UniAP is summarized in the Appendix Algorithm 1.

\noindent\textbf{Graph Initialization.} We use $\mathbf{F}\in R^{HW\times d}$ to denote the L2-normalized feature map with the largest scale from the teacher encoder of the global view, where $H$ and $W$ is the spatial size. To initialize the graph, each token is treated as one node, and edges are formed solely between two nodes that are \textit{directly} adjacent horizontally or vertically. Each $i^{\text{th}}$ node of the $t^{\text{th}}$ layer is associated with a mask $\mathbf{M}_i^t\in\{0, 1\}^{HW}$ denoting which tokens belong to its subtree. We initialize $\mathbf{M}^0$ as identity matrix $\mathbf{I}\in \{0, 1\}^{HW\times HW}$. The initialized \textit{undirected, connected} graph is denoted as $\mathcal{G}^0=\{\mathbf{V}^0, \mathbf{M}^0, \mathbf{E}^0\}$, where $\mathbf{V}^0=\mathbf{F}\in R^{s^0\times d}$ is node features, $\mathbf{E}^0\in \{0, 1\}^{s^0\times s^0}$ is adjacency matrix, $s^0=HW$ is nodes number. We also denote $\mathbf{A}=\mathbf{F}\mathbf{F}^T\in R^{HW\times HW}$ as the spatial affinity matrix, which is of range [-1, 1].

\noindent\textbf{Identify Step.} To identify a group of strongly-connected nodes, for each edge, we first compute the similarity of the two connected nodes. The \textit{feature similarity} of two adjacent $i^{\text{th}}$ and $j^{\text{th}}$ nodes is computed as:
\begin{equation}
    \mathcal{S}^f_{ij}=\mathbf{V}^t_i {\mathbf{V}^t_j}^T\in [-1, 1].
\label{eq:feat_sim}
\end{equation}
Since the node features are evolving at each layer due to the merge step, the pairwise feature similarity may not be consistent with the one implied from the original encoder features. To mitigate this issue, we also measure the \textit{spatial similarity} of two nodes as:
\begin{equation}
\mathcal{S}^s_{ij}=1 - \frac{1}{HW}|\frac{\mathbf{M}^t_i\mathbf{A}}{\mathbf{M}^t_i \mathbf{1}^T} - \frac{\mathbf{M}^t_j \mathbf{A}}{\mathbf{M}^t_j \mathbf{1}^T}| \mathbf{1}^T\in [-1, 1],
\label{eq:spatial_sim}
\end{equation}
where $|\cdot|$ is the absolute operator, $\mathbf{1}\in R^{1\times HW}$ is a vector of 1. According to Equation~\ref{eq:spatial_sim}, if node $i$ and $j$ have similar affinity distribution along the original $HW$ tokens map, their spatial similarity will be large. $\mathcal{S}^s_{ij}$ does not require direct feature comparison, it instead assembles a voting mechanism, in the sense that each original token is a voter which gives its score on $i-j$'s similarity. The final similarity measure is formulated as:
\begin{equation}
    \mathcal{S}_{ij} = \omega_f \mathcal{S}^f_{ij} + \omega_s \mathcal{S}^s_{ij},
\label{eq:final_sim}
\end{equation}
where $\omega_f + \omega_s=1$ and $\mathcal{S}_{ij}\in [-1,1]$. Then, given a threshold $\tau_t$, edges with $\mathcal{S}_{ij}\geq\tau_t$ are labeled to be coarsened. We then use the SCC~\cite{connected_components} algorithm to find groups of strongly-connected nodes. Nodes in each group will be merged into one supernode.

\begin{figure*}[ht]
\begin{center}
\includegraphics[width=\linewidth]{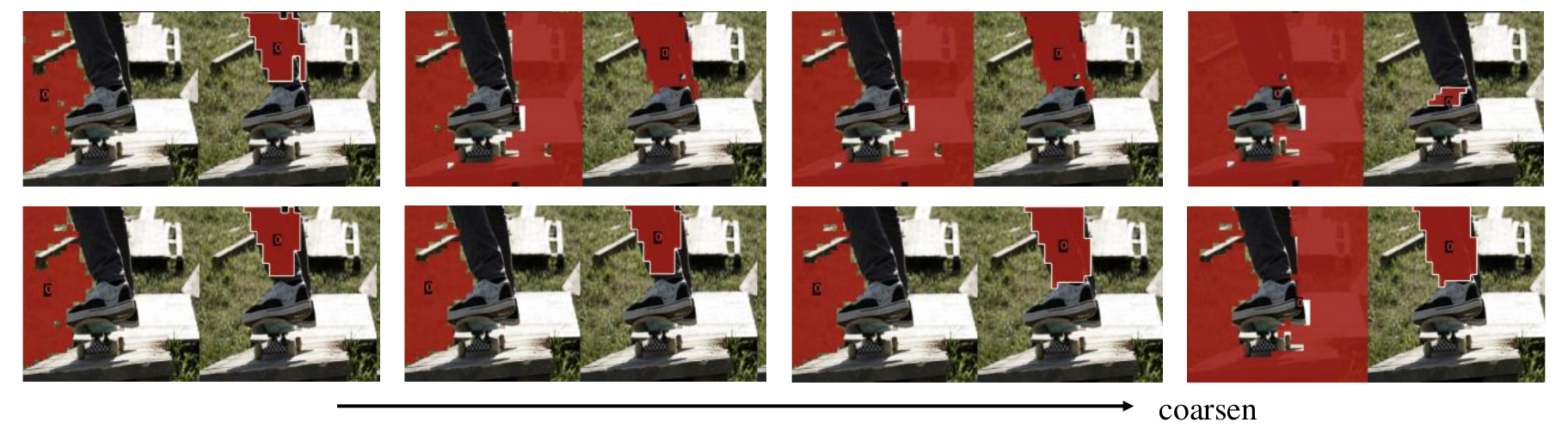}
\end{center}
\vspace{-0.5cm}
\caption{\textbf{Mask visualization of different feature updating strategies.} Each column corresponds to one UniAP layer ($t$=2,3,4,5). (Top) The supernode feature is updated using $\mathbf{V}^{t+1}=\text{L2N}\{\mathbf{\Omega}^T\mathbf{V}^t\}$. (Bottom) The supernode feature is updated using Equation~\ref{eq:feat_update}. Please refer to Appendix C.2 for more qualitative visualizations.}
\label{fig:feat_update}
\end{figure*}

\noindent\textbf{Merge Step.} The SCC algorithm~\cite{connected_components} outputs a node-supernode assignment matrix $\mathbf{\Omega}\in \{0,1\}^{s^t\times s^{t+1}}$. According to the assignment matrix, the new adjacent matrix and mask matrix are updated as:
\begin{equation}
    \mathbf{E}^{t+1}=\mathbf{\Omega}^T\mathbf{E^t} (\mathbf{\Omega}^T\mathbf{E^t})^T; 
    \mathbf{M}^{t+1}=\mathbf{\Omega}^T\mathbf{M}^t,
\label{eq:mask_update}
\end{equation}
where each supernode takes the union of its children nodes masks. To get supernode features, a direct solution is summarizing children features, i.e. $\mathbf{V}^{t+1}=\text{L2N}\{\mathbf{\Omega}^T\mathbf{V}^t\}$. However, as shown in Figure~\ref{fig:feat_update}, we empirically find that this can cause semantically-different but spatially-close nodes to merge earlier. We think this is because the supernode feature updated in this way will treat its boundary tokens and main region tokens equally. When two supernodes are neighboring to each other, their common boundary tokens will make their features have a larger similarity. To mitigate this issue, the supernode feature is computed as:
\begin{equation}
    \mathbf{V}^{t+1}_i=\text{L2N}\{\text{softmax}(\frac{\mathbf{M}^{t+1}_i\mathbf{A}}{\sigma})\mathbf{F}\},i=1,...,s^{t+1}
\label{eq:feat_update}
\end{equation}
where $\text{L2N}\{\cdot\}$ denotes L2-normalization, $\sigma$ is the softmax temperature. Equation~\ref{eq:feat_update} can also be seen as a weighted voting mechanism, where most of the softmax mass will lie on the main region of supernode, and information of boundary tokens is suppressed. As shown in Figure~\ref{fig:feat_update}, the ``grass'' region are not merged with ``human leg'' region throughout all layers. This shows that updating supernode features using Equation~\ref{eq:feat_update} can make semantically-different but spatially-close regions more discriminative.

\noindent\textbf{Universal Pseudo-masks Generation with Instance Pooling and Semantic Pooling.} It is noteworthy that edges exist only between directly adjacent nodes in $\mathcal{G}^0$. Moreover, Equation~\ref{eq:mask_update} ensures edges are maintained only between adjacent supernodes, i.e. sparsely connected. Consequently, each mask corresponds to a consolidated region. It is sensible for instance segmentation but not for semantic segmentation, which requires masks that encompass multiple disjoint regions belonging to the same semantic class. As shown in Figure~\ref{fig:overview}, to address this limitation, we construct a \textit{fully connected} version of the graph $\mathcal{G}^*$ derived from instance pooling. Subsequently, the same Identify and Merge pipeline is employed to generate semantic masks, thereby combining multiple disjoint components into a single semantic-level mask. The graph derived from instance pooling is used as input of the next UniAP layer.

\noindent\textbf{Multi-Granular Pseudo-masks Generation with Time-varied Thresholding.} UniAP can be seen as an \textit{iterative} coarsening procedure. After each iteration, each supernode will represent its receptive region defined by $\mathbf{M}_i^t$. By defining a set of decreasing thresholds $\{\tau_t|\tau_{t+1}<\tau_t\}_{t=1}^{\Gamma}$, we can  get masks at different hierarchy. We empirically find that background tokens, where pixels are very similar to their neighbors, are merged at the very early layers, i.e. with much higher threshold close to 0.9. While tokens like ``human head'' and ``human body'' each with higher level semantics are merged at later layers with a lower threshold close to 0.5. Readers can refer to Appendix C.2, Appendix Figure 5, and Appendix Figure 6 for more detailed visualizations.

\subsection{Model Training and Initialization}
\label{sec:training}
To deploy UniAP for scalable single-stage pretraining, we devise a teacher-student framework with momentum updating strategy.  Moreover, we introduce a novel pretext task, Query-wise Self-Distillation, that self-distills in a query-wise manner to facilitate segmentation-oriented self-supervised learning.

\noindent\textbf{Multi-scale encoder and zero-initialization.} Existing self-supervised models~\cite{cutler, unsam, u2seg} utilize ResNet~\cite{resnet} as backbone, whereas they use DINO~\cite{dino} pretrained ViT-base/8~\cite{vit} to generate pseudo-masks. Our framework only devises ViT-base/8 as backbone. We also follow ViT-Adapter~\cite{vit_adapter} to augment ViT with lightweight multi-scale adapters. To save computation cost, we scale down the image input to the ViT branch by half as if using a patch size of 16. To generate sensible pseudo-masks as training begins, we zero-initialize the adapters so that UniAP can utilize the pretrained DINO features at the very start.

\noindent\textbf{Task-specific Object Queries.} Unlike U2Seg~\cite{u2seg} where the decoder has two separate branches, we adopt a more streamlined approach by partitioning the object queries in mask decoder into semantic queries and instance queries with two special learnable tokens \texttt{[INS]} and \texttt{[SEM]}. It is noteworthy that both groups of queries share the same decoder parameters and projection head, enabling universal segmentation through parameter sharing.

\noindent\textbf{Cropped Bipartite Matching.} The teacher branch processes the original \textit{global} view of the image, producing pseudo-masks of the entire image. In contrast, the student branch processes \textit{local} views obtained from multi-crop augmentation. For each local view, we crop the teacher's pseudo-masks correspondingly to match the student's view position. Teacher queries and pseudo-masks that do not overlap with the student's view are dropped. Only mask Dice similarity is used as the matching criterion. The matching processes for the semantic and instance levels are independent and do not interfere with each other.

\noindent\textbf{Query-wise Self-Distillation.} We denote the query embeddings of the teacher and student branch after bipartite matching as $\mathbf{Q}_t\in R^{L\times d}$ and $\mathbf{Q}_s\in R^{L\times d}$, where $L$ is the query number. A projection head of three-layer MLP is used to transform queries into distribution logits. The Query-wise Self-distillation (QuerySD) pretext is formulated as:
\begin{equation}
    \sum_{t=1, s\in\{1,...,\delta\}}\sum_{l=1}^{L} \sum_{k=1}^{d}\mathbf{Q}_t^{l,k}\text{log}\mathbf{Q}_s^{l,k},
\label{eq:query_distill}
\end{equation}
where $\delta$ is the number of local crops.
Loss~\ref{eq:query_distill} can be interpreted as the sum of self-distillation loss~\cite{dino} over each pair of matched teacher query and student query. 
Unlike other self-supervised \textit{representation} frameworks which use the ViT \texttt{[CLS]} token or ResNet global pooling, self-supervised \textit{segmentation} requires more fine-grained features for pretext training. 
S2-UniSeg first aggregates each image into a set of object queries, and self-distills in a query-wise manner to facilitate segmentation-oriented self-supervised learning.
\begin{table*}[htb]
    \small
    \centering
    \resizebox{\linewidth}{!}{
    \begin{tabular}{l|cc|cc|cc|cc|ccc|ccc}
    \toprule
    Task       & \multicolumn{2}{c|}{ \makecell[c]{Class-agnostic\\Instance Seg.}} & \multicolumn{4}{c|}{Instance Seg.}  & \multicolumn{2}{c|}{Semantic Seg.} & \multicolumn{6}{c}{Panoptic Seg.}     \\ \hline
    Datasets & \multicolumn{2}{c|}{COCO}                  & \multicolumn{2}{c|}{COCO} & \multicolumn{2}{c|}{UVO} & \multicolumn{2}{c|}{COCOStuff-27}         & \multicolumn{3}{c|}{COCO} & \multicolumn{3}{c}{Cityscapes} \\ \hline
    Metric   &   AP     &      AR                 & AP         & AR         & AP         & AR       & PixelAcc          & mIoU         & PQ     & SQ     & RQ     & PQ       & SQ       & RQ      \\ \hline
    STEGO~\cite{stego} (COCOStuff)     & -                   & -                   & -           & -          & -          & -          & 56.9              & 28.2         & -     & -      & -      & -        & -        & -       \\ 
    CutLER~\cite{cutler} (IN)    & 9.7                & 27.1                & -           & -          & -          & -          & -                 & -            & -      & -      & -      & -        & -        & -       \\
    U2Seg~\cite{u2seg} (IN)     & 7.3             & 19.2               & 6.4     & 18.5   & 6.2       & 21.0       & 63.9             & 30.2         &16.1   & 71.1   & 19.9   & 17.6    & 52.7     & 21.7    \\ 
    \rowcolor{gray!10}
    S2-UniSeg (IN)     & \textbf{14.2}               & \textbf{34.3}                & \textbf{15.3}        & \textbf{35.5}   & \textbf{16.2}       & \textbf{32.1}       & \textbf{68.4}              & \textbf{36.1}         & \textbf{20.2}   & \textbf{80.6}   & \textbf{26.7}   & \textbf{20.5}     & \textbf{60.1}     & \textbf{29.7}    \\  \hline 
    UNSAM~\cite{u2seg} (SA-1B 0.4M)    & 31.4                & 42.0                & -        & -          & -          & -          & -                 & -            & -      & -      & -      & -        & -        & -       \\
    \rowcolor{gray!10}
    S2-UniSeg (SA-1B 0.4M)     & \textbf{33.6}               & \textbf{43.9}                & \textbf{27.4}        & \textbf{36.1}  & \textbf{20.8}       & \textbf{34.6}       & \textbf{72.3}              & \textbf{37.2}         & \textbf{25.7}   & \textbf{88.4}   & \textbf{28.6}   & \textbf{22.5}     & \textbf{63.7}     & \textbf{32.6}    \\ 
    \rowcolor{gray!10}
    S2-UniSeg (SA-1B 1.2M)     & \textbf{35.1}               & \textbf{44.6}                & \textbf{29.1}        & \textbf{38.4}   & \textbf{22.6}       & \textbf{35.7}       & \textbf{74.4}              & \textbf{38.9}         & \textbf{28.3}   & \textbf{89.2}   & \textbf{29.3}   & \textbf{23.6}     & \textbf{64.3}     & \textbf{33.4}    \\  
    \rowcolor{gray!10}    
    S2-UniSeg (SA-1B 2.0M)     & \textbf{36.4}               & \textbf{46.3}                & \textbf{30.7}        & \textbf{40.3}   & \textbf{24.3}       & \textbf{36.4}       & \textbf{76.8}              & \textbf{39.7}         & \textbf{29.6}   & \textbf{90.3}   & \textbf{30.2}   & \textbf{25.4}     & \textbf{66.7}     & \textbf{35.2}    \\  \hline
    \end{tabular}
    }
    \caption{Performance comparison with previous methods on zero-shot class-agnostic instance segmentation, unsupervised instance-segmentation, unsupervised semantic segmentation, and unsupervised panoptic segmentation. Our model outperforms other state-of-the-art methods on all tasks by a large margin. Moreover, pretrained on the large-scale 2M subset from SA-1B~\cite{sam}, S2-UniSeg achieves significant improvements on class-agnostic instance segmentation compared with UnSAM~\cite{unsam}. Please refer to Appendix B and Appendix C for full results and more qualitative visualizations.}
    \label{tab:comp_full}
\end{table*}

\section{Experiments}
\label{sec:experiments}

\subsection{Training Settings}
\textbf{Model Architecture.} Following previous works~\cite{u2seg, cutler, unsam}, we use the DINO pretrained ViT-base/8 and ViT-Adapter~\cite{vit_adapter} as backbone. For the mask decoder, we use the official setting of Mask2Former~\cite{mask2former}. For the project head, we follow DINO~\cite{dino} using a 3-layer MLP with hidden dimension 2048 followed by L2 normalization and a linear layer of $K$ dimensions.
After training, the final teacher encoder with the mask decoder is used for inference. \noindent\textbf{Hyper-parameter setting.} We set $\sigma=0.07, \{\tau_t\}_{t=1}^{\Gamma}=[0.8, 0.7. 0.6, 0.5, 0.4], \phi=5, \omega_f=0.6, \omega_s=0.4, K=512$. The number of semantic and instance queries are set to 50 and 150. A local multi-crop scale between 0.05 and 0.4 is used. \noindent\textbf{Optimization setting.} We train the model using AdamW with a batch size of 16. The learning rate is linearly warmed up to 0.000625 for 10k iterations. Our model is trained for 160k iterations, while other CutLER models are additionally self-trained for several 80K iterations.

\subsection{Datasets and Metrics Details}
\label{sec:app_data}
Following U2Seg~\cite{u2seg} and UnSAM~\cite{unsam}, we test unsupervised instance segmentation on COCO\texttt{val}~\cite{coco}, and UVO\texttt{val}~\cite{uvo}. We test unsupervised semantic segmentation on COCOStuff-27~\cite{cocostuff}. Following U2Seg ~\cite{u2seg}, we test unsupervised panoptic segmentation on Cityscapes\texttt{val}~\cite{cityscape} and COCO \texttt{val}. Following U2Seg, the pseudo-classes are mapped using Hungarian matching to class labels. We use the ImageNet-1k (1.3M images) dataset for pretraining. Please refer to Appendix D for more dataset details 

\subsection{Experiment Results}
\begin{figure*}[ht]
\begin{center}
\includegraphics[width=0.8\linewidth]{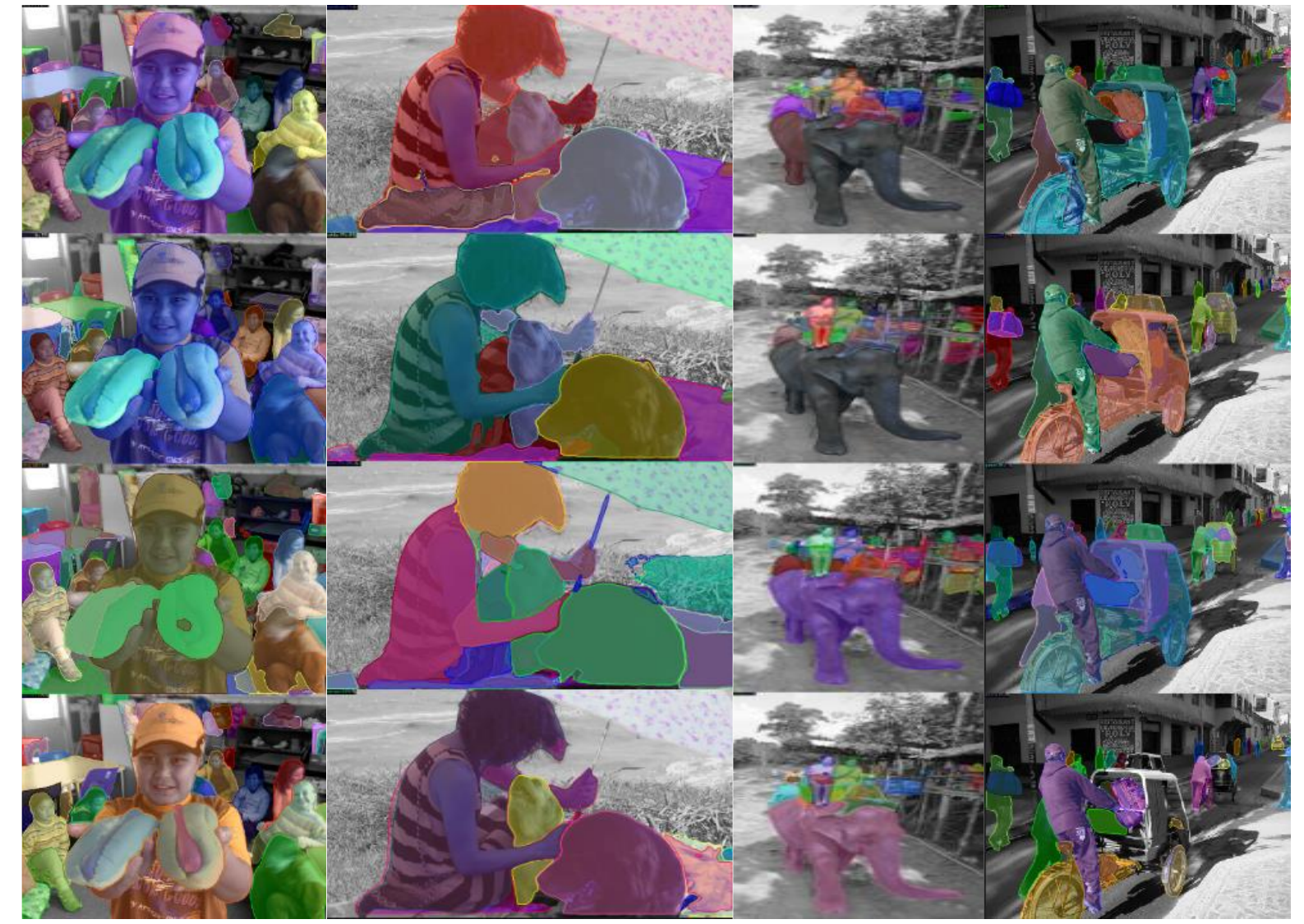}
\end{center}
\caption{Comparison between U2Seg~\cite{u2seg}, CutLER~\cite{cutler}, UnSAM~\cite{unsam}, and S2-UniSeg (ours) on Unsupervised Class-agnostic Instance Segmentation. Each row from top to botton is U2Seg, CutLER, UnSAM, and S2-UniSeg. Please refer to Appendix B and Appendix C for more comparisons.}
\label{fig:ins2}
\end{figure*}

\textbf{Self-supervised Instance Segmentation.}
As shown in Table~\ref{tab:comp_full}, for class-agnostic unsupervised instance segmentation, S2-UniSeg achieves an increase of \textbf{+6.9} in AP and \textbf{+15.1} in AR, which is 94.5\% and 78.6\% increase compared to U2Seg~\cite{u2seg}. As shown in Figure~\ref{fig:comparision}, our online pretraining framework can converge faster and achieve significant improvements over previous multi-stage alternating methods. Moreover, for our method, the performance of the class-aware instance segmentation is higher of that in class-agnostic instance segmentation. However, this is reversed for U2Seg. This shows that our online clustering QuerySD pretext task can better help model to capture the semantics information in the pretraining dataset. 


\noindent\textbf{Self-supervised Semantic Segmentation.}
As shown in Table~\ref{tab:comp_full}, S2-UniSeg outperforms other state-of-the-art methods on COCOStuff-27 for unsupervised semantic segmentation. Specifically, our model achieves an increase of \textbf{+5.9} in PixelAcc, which is 19.5\% of increase compared to U2Seg and 20.9\% of increase compared to STEGO. Please refer to Appendix B and Appendix C for more qualitative comparisons.

\noindent\textbf{Self-supervised Panoptic Segmentation.}
As shown in Table~\ref{tab:comp_full}, S2-UniSeg also significantly outperforms U2Seg over panoptic segmentation on COCO and Cityscapes. For the zero-shot setting (solely trained on ImageNet), our method achieves an increase of \textbf{+6.8} in SQ on Cityscapes, which is 14.5\% of increase compared to U2Seg, and achieves an increase of \textbf{+5.1} in PQ on COCO, which is 45.9\% of increase compared to U2Seg. Please refer to Appendix B and Appendix C for more qualitative comparisons.

\subsection{Scaling S2-UniSeg with SA-1B.}
Since S2-UniSeg is a single-stage pretraining framework without any time-consuming offline processes, we can easily scale S2-UniSeg with larger dataset size. We pretrain S2-UniSeg on subsets of SA-1B with different scale, i.e. 0.4M, 1.2M, and 2M for 160K iterations. The optimization landscape is plotted in Figure~\ref{fig:comparision}. Compared with the state-of-the-art UnSAM~\cite{unsam} framework which adopts the discontinuous CutLER routine, S2-UniSeg effectively scales with dataset size. For class-agnostic instance segmentation, S2-UniSeg achieves \textbf{+2.4} AP and \textbf{+1.9} AR using the same amount of data (0.4M). The 2M pretrained version achieves the highest performance with \textbf{46.3} AR and \textbf{36.4} AP. Moreover, the SA-1B pretrained S2-UniSeg also sets new record on other three benchmarks.


\subsection{Ablation Studies}
We identify four main hyper-parameters for ablation studies, which are the spatial and feature similarity weights ($\omega_f, \omega_s$), UniAP threshold schedule $\{\tau_t\}_{t=1}^{T}$, the number of QuerySD projection head dimension, and the local crops number. We evaluate all ablations on the unsupervised class-aware instance segmentation on UVO\texttt{val} dataset.

\begin{table}[t]
    \small
    \centering
    \resizebox{0.47\linewidth}{!}{%
    \begin{tabular}{ccc}
    \toprule
    \makecell[c]{Similarity weights\\$\omega_s, \omega_f$}  &  AP$^\text{mask}$ & AR$^\text{mask}_{100}$ \\ \hline
    (0.6, 0.4)   & 15.1 & 27.3     \\
    (0.5, 0.5)   & 15.7 & 28.4     \\ 
    (0.4, 0.6)   & 16.2 & 32.1      \\ 
    (0.0, 1.0)   & 12.4 & 25.4      \\ \toprule
    \end{tabular}}
    \resizebox{0.45\linewidth}{!}{%
    \begin{tabular}{ccc}
    \toprule
    \makecell[c]{UniAP thresholds\\ $\{\tau_t\}_{t=1}^{T}$}     &  AP$^\text{mask}$ & AR$^\text{mask}_{100}$ \\ \hline
    (0.9-0.1)   & 16.7 & 31.8     \\
    (0.5-0.1)   & 15.5 & 26.3      \\ 
    (0.8-0.4)   & 16.2 & 32.1   \\ \toprule
    \end{tabular}}
    \resizebox{0.45\linewidth}{!}{%
    \begin{tabular}{ccc}
    \toprule
   \makecell[c]{Local crops\\ number $\delta$  } &  AP$^\text{mask}$ & AR$^\text{mask}_{100}$ \\ \hline
    2   & 16.2 & 32.1     \\
    4   & 15.9 & 31.6      \\ 
    6   & 16.0 & 32.3  \\ \toprule
    \end{tabular}}
    \resizebox{0.47\linewidth}{!}{
       \begin{tabular}{ccc}
    \toprule
    \makecell[c]{Projection \\ Dimension $K$  } &  AP$^\text{mask}$ & AR$^\text{mask}_{100}$ \\ \hline
    128   & 13.6 & 27.4     \\
    512    & 16.2 & 32.1     \\  
    1024   & 16.9 & 33.2      \\ \toprule
    \end{tabular}}
    \caption{Ablation studies for the spatial and feature similarity weights, threshold schedule, local crops number, and QuerySD projection dimension.}
    \label{tab:ablation}
\end{table}

\noindent\textbf{Feature and spatial similarity weights.} As shown in Table.~\ref{tab:ablation}, without considering spatial similarity, i.e. setting $\omega_s=0$, the performance drops significantly. By setting $\omega_s=0.4$, our model achieves an increase of \textbf{+3.8} AP. This validates our design for the Identify step.

\noindent\textbf{Time-varied UniAP thresholds.} As shown in Table.~\ref{tab:ablation}, when we use a set of much lower thresholds, the performance drops significantly. This is because a lower threshold (0.5) will make almost every edge to be coarsened. Instead, by using a more fine-grained set of thresholds (0.9,0.8,...,0.1), the UniAP layer can identify more fine-grained groups of different semantic hierarchies. However, this would cost much time since more UniAP layers are used. Instead, by setting an intermediate set of thresholds (0.8,...,0.4), our model can have comparable performance and also cost less time.

\noindent\textbf{Local crops number.} As shown in Table.~\ref{tab:ablation}, different from DINO~\cite{dino} which shows sensitive to multi-crops number. S2-UniSeg is more robust. Since more local crops mean more bipartite matching and each matching costs considerable time, we just crop 2 local views to save computation.

\noindent\textbf{QuerySD projection head dimension $K$.} As shown in 
 Table.~\ref{tab:ablation}, by using more clusters, our model can learn a finer representation granularity.  This is also validated in U2Seg and previous self-supervised representation models. However, unlike DINO where a large number of 65536 is used, we find a smaller $K$ is adequate for satisfactory self-supervised segmentation performance.

\section{Limitation}
While the UniAP algorithm demonstrates promising results, it involves a number of hyperparameters that may necessitate additional tuning. Although the current configuration performs effectively across various scenarios, further refinement of these parameters could improve its performance and expand its applicability to a broader set of use cases.
\section{Conclusion}
\label{sec: conclusion}
In this paper, we propose an efficient pseudo-mask generation algorithm, Fast Universal Agglomerative Pooling (UniAP), to generate both universal and multi-granular masks of one image within tens of milliseconds. We also propose a novel segmentation-oriented pretext task, Query-wise Self-Distillation, to train a student and a momentum teacher with single-stage online pretraining. S2-UniSeg achieves state-of-the-art performance on unsupervised zero-shot instance segmentation, semantic segmentation, and panoptic segmentation tasks. Moreover, S2-UniSeg demonstrates strong scalability with increasing training data, yielding better performance as the dataset size grows.

\bigskip

\bibliography{aaai2026}


\clearpage

\newpage
\appendix
Our supplementary materials include additional visual and qualitative comparisons in Section C, more detailed information about the datasets in Section D, further training details, details of post-processing bipartite matching for  class-aware semantic, instance, and panoptic segmentation in Section F,  and the complete quantitative results across 3 benchmarks in Section G. Please refer to the corresponding section for further details and clarifications of each part.
\section{A. Preliminaries}
\label{sec:preli}
\textbf{Unsupervised Universal Image Segmentation (U2Seg).} Universal segmentation requires annotations of instance-level masks (``thing''), semantic-level masks (``stuff``), and the class label for each mask. As the first self-supervised universal segmentor, U2Seg~\cite{u2seg} follows the cut-and-learn~\cite{cutler} pretraining pipeline. The ``cut'' stage is offline and generates universal pseudo-masks for the whole dataset. Specifically, based on SSL feature maps~\cite{dino}, U2Seg uses MaskCut~\cite{cutler} to generate class-agnostic instance-level pseudo-masks for each image. K-means is then used to cluster these masks and get their pseudo-labels. Moreover, U2Seg directly uses the pretrained model STEGO~\cite{stego} to get the semantic segmentation pseudo-masks. The ``learn'' stage alternates between training model using the current annotations and generating new pseudo annotations using the previous model checkpoint. 

\noindent\textbf{Self-distillation with No Labels (DINO).}
Self-supervised representation learning often incorporates a pretext task to supervise encoder training. DINO~\cite{dino} is a self-distillation model that leverages online clustering~\cite{swav, sela} with multi-crop invariance learning~\cite{swav}. Specifically, DINO employs a student and momentum teacher~\cite{moco}, wherein both networks consist of an encoder and a projection head. During training, an unlabeled image is first cropped into 2 bigger global-views and 8 smaller local-view, the teacher encodes solely global views, whereas the student encodes all views. ViT~\cite{vit}\texttt{[CLS]} token or ResNet~\cite{resnet} global pooling is then input to the projection head. The DINO self-distillation pretext is formulated as:
\begin{equation}
   \sum_{t\in\{1,2\}, s\in\{1,...,10\}, s\neq t} \hspace{-8mm}SD(\mathbf{p}_t,\mathbf{p}_s) = -\sum \sum_{k=1}^{K}\mathbf{p}_t^k \text{log} \mathbf{p}_s^k,
\label{eq:dino_loss}
\end{equation}
where $K$ is the online cluster number (head output dimension), $\mathbf{p}_t\in R^K$ is softmax normalized teacher head output, $\mathbf{p}_s\in R^K$ is softmax normalized student head output. This alignment from local view to global view enables the model to capture local-to-global semantic invariance. The teacher network is updated using a momentum mechanism~\cite{moco} that effectively ensembles the model over time~\cite{mean_teacher}, thus providing better global features to guide the student and enhance learning without the need for manual labels.

\section{B. Detailed algorithm of UniAP}
Please refer to Algorithm 1 for details.
\begin{algorithm*}[ht]
    \caption{Fast Universal Agglomerative Pooling}\label{alg:algorithm}  
\begin{algorithmic}[1]
\Require{{Initialized graph: $\mathcal{G}^0=\{\mathbf{V}^0,\mathbf{M}^0,\mathbf{E}^0\}$},
         {Spatial affinity matrix: $\mathbf{A}\in R^{HW\times HW}$},
         {L2-normalized key features: $\mathbf{F}\in R^{HW\times d}$}, 
         {Time-varied thresholds: $\{\tau_t\}_{t=1}^{T}$}, 
          {Softmax temperature: $\sigma$}, {Spatial and Feature similarity weight: $\omega_s, \omega_f$}, {Mask area threshold $\phi$}.}
    \vspace{0.1cm}\hrule\vspace{0.1cm}
    \State $\textit{instance masks}\gets\{\}$, $\textit{instance features}\gets\{\}$, $\textit{semantic masks} \gets \{\}$, $\textit{semantic features}\gets \{\}$
    \For{$t$ \textbf{in} $[0, ..., \Gamma-1]$}
        \State \text{Compute similarity score of each edge as Equation~\ref{eq:final_sim}}.
        \State $\text{Find set of edges }\{E^t_i\}\text{ with score larger than }\tau_{t+1}$.
        \State $\text{Compute assignment matrix }\mathbf{\Omega}=\text{SCC}(\{E^t_i\})$.
        \State $\text{Update }\mathbf{E}^{t+1}\gets \mathbf{\Omega}^T\mathbf{E^t} (\mathbf{\Omega}^T\mathbf{E^t})^T$.
        \State $\text{Update }\mathbf{M}^{t+1}\gets\mathbf{\Omega}^T\mathbf{M}^t$.
        \State $\text{Repeat steps 3-7 until all edges' score }<\tau_{t+1}$.
        \For{$i$ \textbf{in} $[0, ..., s^{t+1}-1]$}
            \State $\text{Update }\mathbf{V}^{t+1}_i\gets \text{L2N}\{\text{softmax}(\frac{\mathbf{M}^{t+1}_i\mathbf{A}}{\sigma})\mathbf{F}\}$.
            \If{$area(\mathbf{M}^{t+1}_i)\geq\phi$}
                \State $\text{Append }\mathbf{M}^{t+1}_i\text{ to } \textit{instance masks}$. 
                \State $\text{Append }\mathbf{V}^{t+1}_i\text{ to } \textit{instance features}$.
            \EndIf
         \EndFor
         \State $\text{Build fully-connected }\mathcal{G}^*=\{\mathbf{V}^{t+1},\mathbf{M}^{t+1},\mathbf{E}^*\}$.
         \State $\text{Repeat steps 3-15, get \textit{semantic masks \& features}}$.
    \EndFor
\end{algorithmic}
\end{algorithm*}

\section{C. Visualization and Qualitative Comparisons}
\label{sec:visual}

\subsection{C.1 Qualitative comparisons between S2-UniSeg and STEGO~\cite{stego}, U2Seg~\cite{u2seg}, CutLER~\cite{cutler}, and UnSAM~\cite{unsam}.}
\label{sec:visual1}

We provide more visual comparisons between STEGO, U2Seg, CutLER, UnSAM and our method on unsupervised class-agnostic instance segmentation (Figure~\ref{fig:ins}), unsupervised panoptic segmentation (Figure~\ref{fig:pano}), and unsupervised instance segmentation (Figure~\ref{fig:instace}), unsupervised semantic segmentation (Figure~\ref{fig:sem}).

\subsection{C.2 Interpretable visualizations of each UniAP layer}
\label{sec:inside_uniap}

\subsubsection{C.2.1 Pooling process of each UniAP layer.}
\clearpage
\begin{figure*}[p]
\begin{center}
\includegraphics[width=\linewidth]{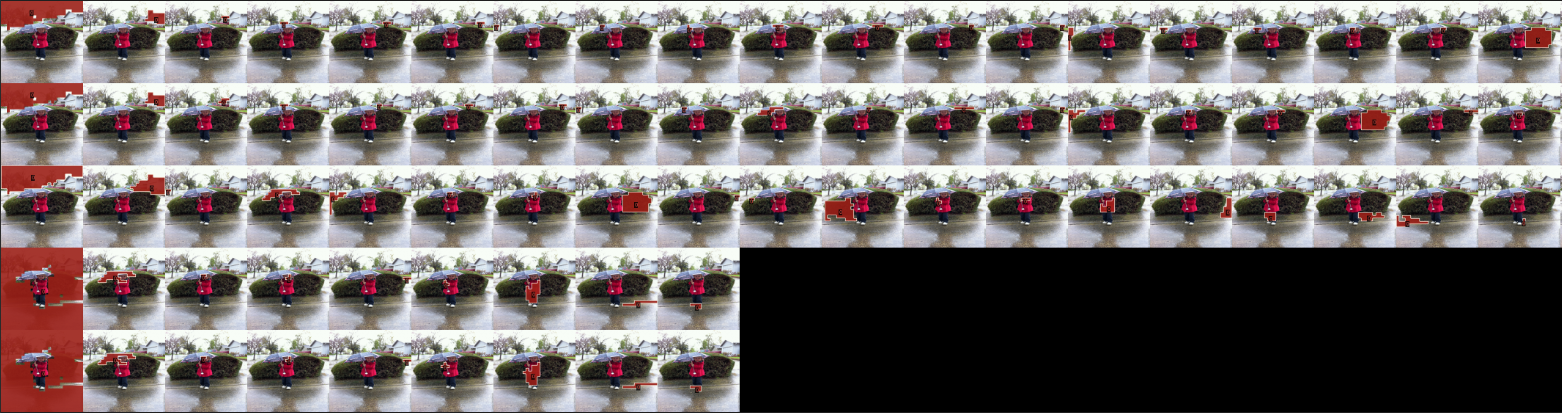}
\includegraphics[width=\linewidth]{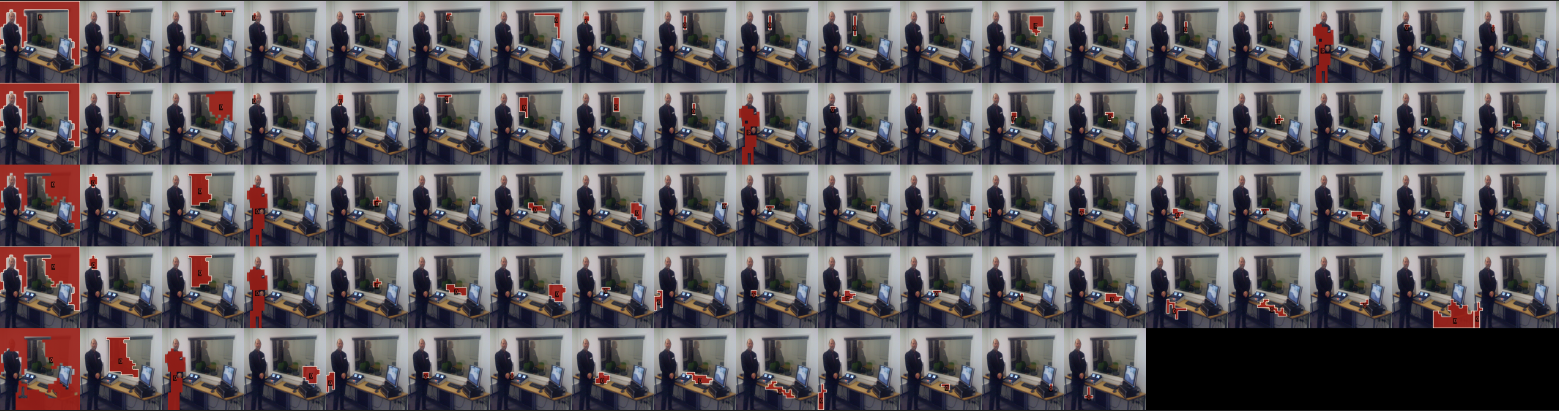}
\includegraphics[width=\linewidth]{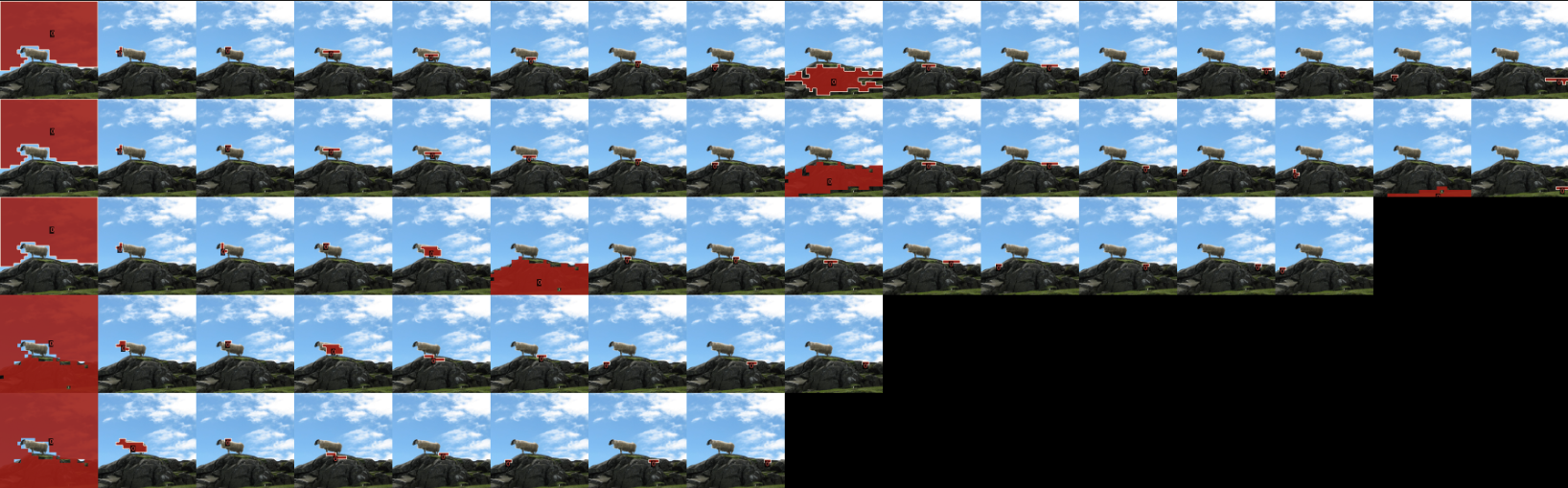}
\includegraphics[width=\linewidth]{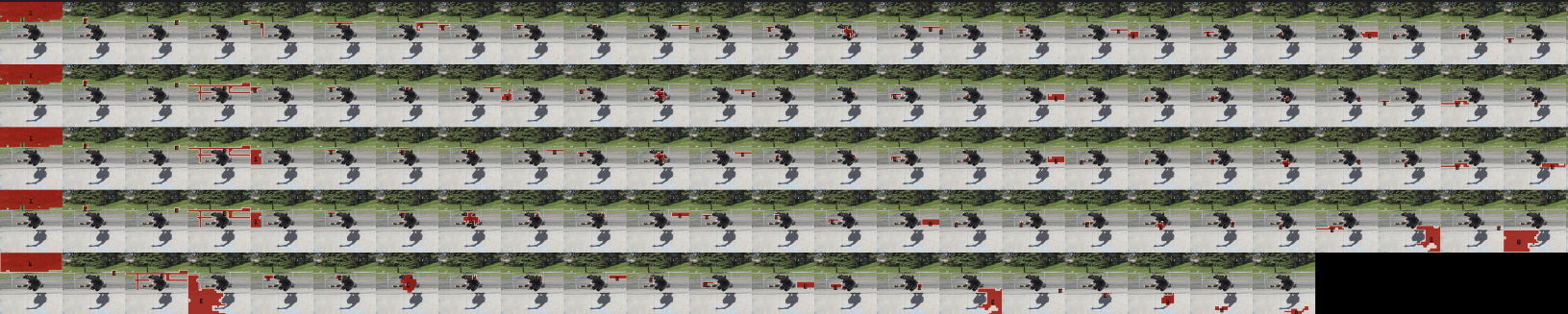}
\end{center}
\caption{\textbf{Four visualizations of the pooling process of each UniAP layer (a total of 5 layers from top to bottom).} Since each layer has tens of nodes, we only visualize some part of the pooled nodes at earlier layers.}
\label{fig:v2}
\end{figure*}
\clearpage
As shown in Figure~\ref{fig:v2}, we visualize the pooling process controlled by the time-varied thresholding procedure of each UniAP layer. Each row is a visualization of each UniAP layer. Due to row space limitation, we have to visualize only a partial set of nodes at earlier layers.

We can see that, at earlier layers, since we use a larger threshold (0.8), \textit{neighboring} visually-similar regions, such as sky, grass, wall are firstly clustered. While for regions like the whole person which is often composed of many different regions (leg, head, pole), they are clustered at later UniAP layers. 

\subsubsection{C.2.2 Visual attentions of each UniAP layer.}
\begin{figure*}[htb]
\begin{center}
\includegraphics[width=\linewidth]{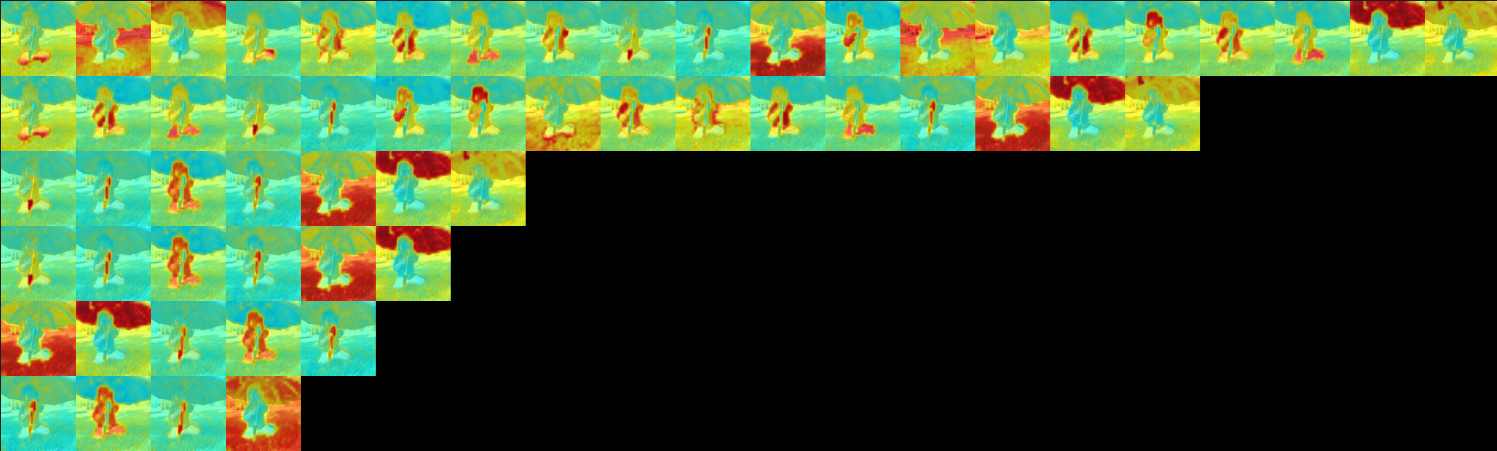}
\end{center}
\caption{\textbf{The node attention heatmap of each UniAP layer (a total of 5 layers.)}}
\label{fig:v1}
\end{figure*}
As shown in Figure~\ref{fig:v1}, we visualize the attention heatmap of all nodes in each UniAP layer. The heatmap is computed using the inner product of the node feature with the image features. It should be noted that we prune nodes that share very similar pseudo-masks (i.e. Non Maximum Suppression (NMS) on the set of pseudo-masks of all intermediate generated nodes at all UniAP layers).

We can see that, at earlier layers, the UniAP would more likely to aggregate the background tokens. This is consistent with our intuition that background tokens are visually-similar and have abundant information, therefore their DINO~\cite{dino} pretrained features will have a larger cosine similarity on the radius-1 sphere space learned by the DINO representation.  At final layer, some regions with higher-level semantics are clustered together, such as the umbrella pole, the girl, and the background.

\begin{figure*}[htb]
\begin{center}
\includegraphics[width=\linewidth]{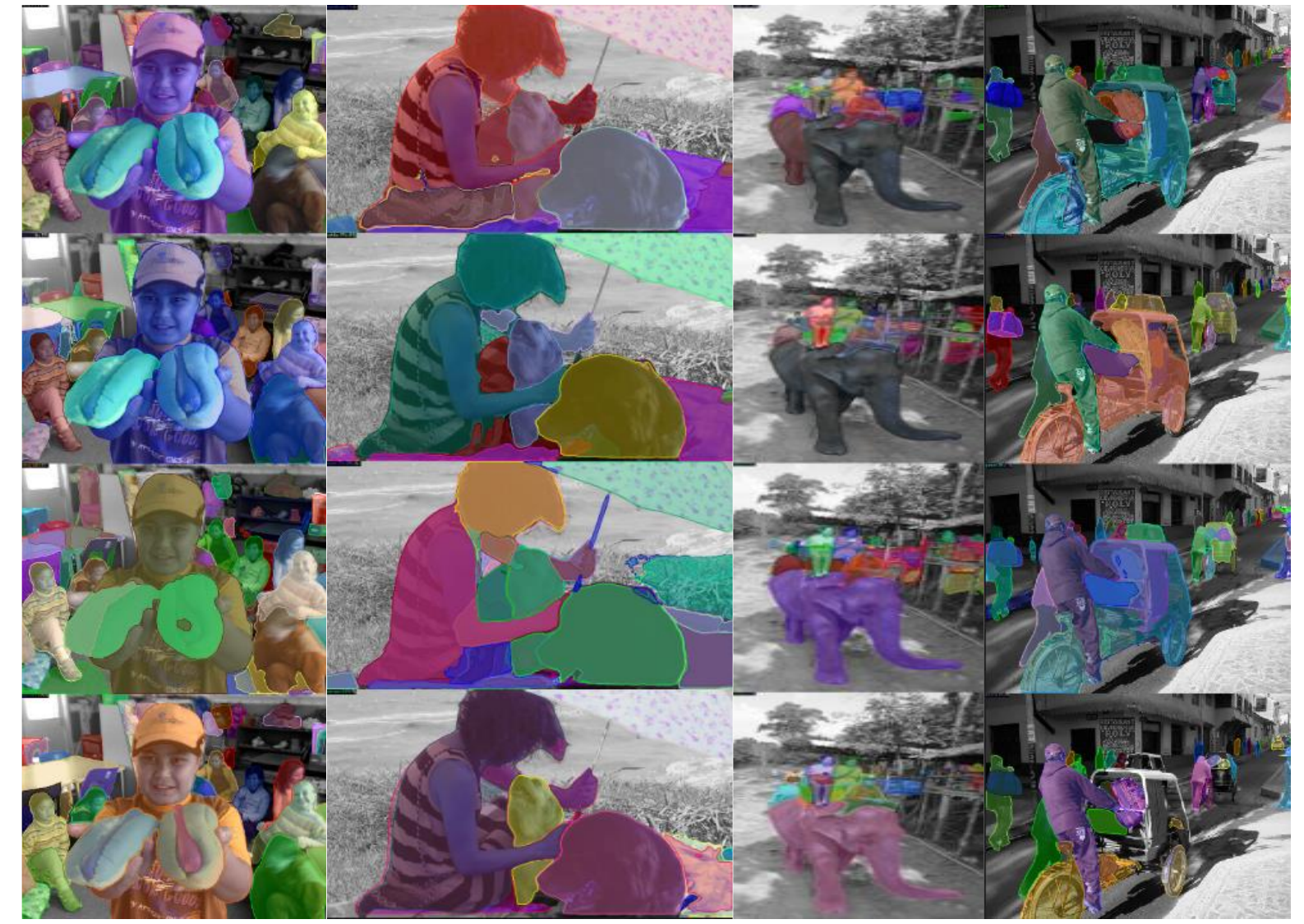}
\end{center}
\caption{\textbf{Comparison between U2Seg~\cite{u2seg}, CutLER~\cite{cutler}, UnSAM~\cite{unsam}, and S2-UniSeg (ours) on Unsupervised Class-agnostic Instance Segmentation.} Each row from top to botton is U2Seg, CutLER, UnSAM, and S2-UniSeg}
\label{fig:ins}
\end{figure*}

\begin{figure*}[htb]
\begin{center}
\includegraphics[width=\linewidth]{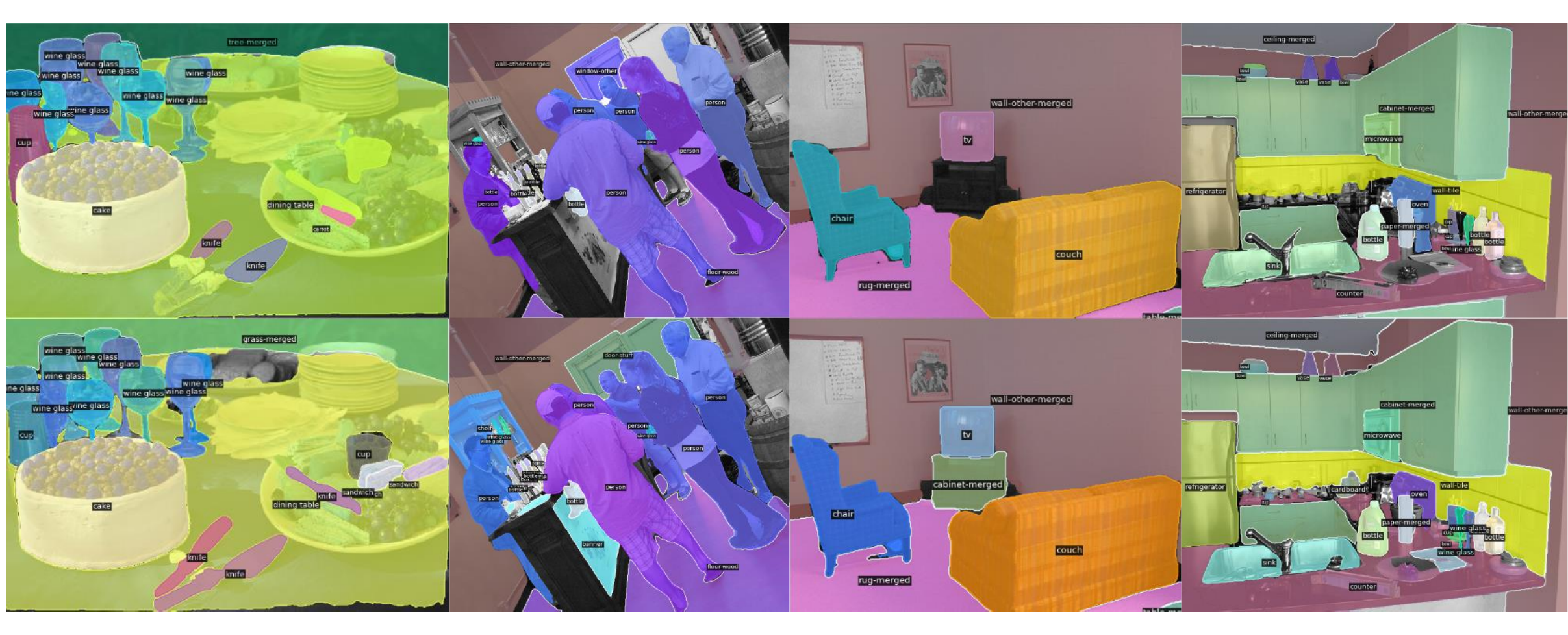}
\end{center}
\caption{\textbf{Comparison between U2Seg and S2-UniSeg on Unsupervised Panoptic Segmentation.} (top) Visualizations of Panoptic Segmentation of U2Seg. (bottom) Visualizations of Panoptic Segmentation of S2-UniSeg.}
\label{fig:pano}
\end{figure*}

\begin{figure*}[htb]
\begin{center}
\includegraphics[width=\linewidth]{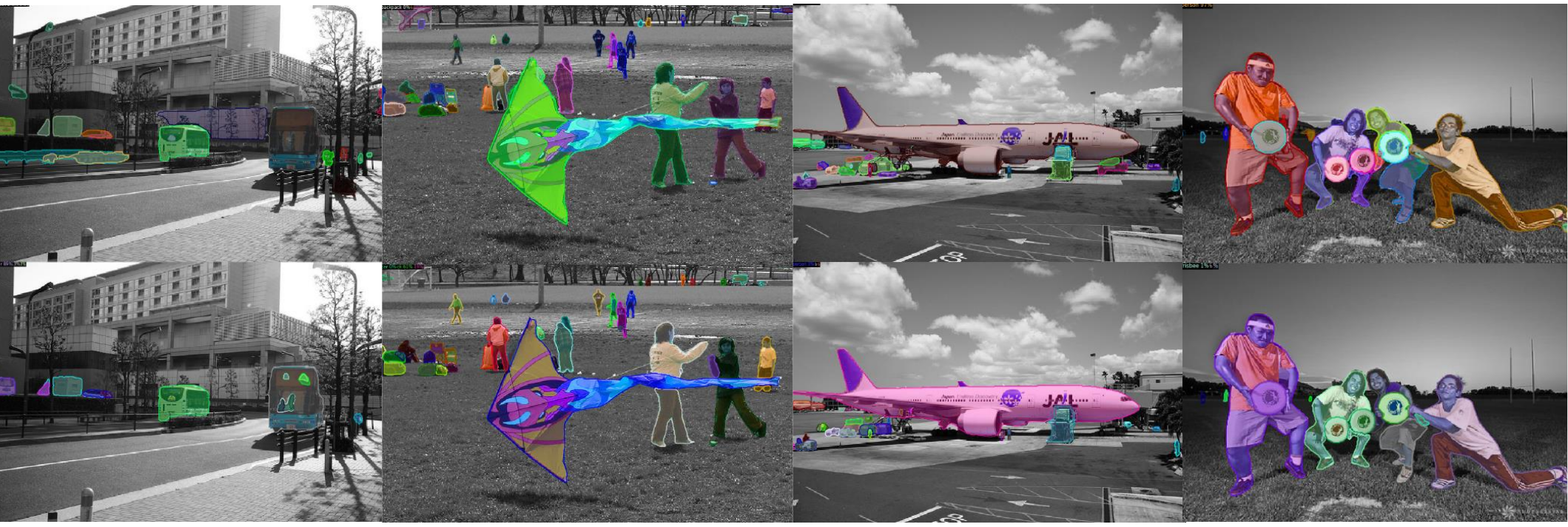}
\end{center}
\caption{\textbf{Comparison between U2Seg, and S2-UniSeg on Unsupervised Instance Segmentation.} (top) Visualizations of Instnace Segmentation of U2Seg. (bottom) Visualizations of Instance Segmentation of S2-UniSeg.}
\label{fig:instace}
\end{figure*}

\begin{figure*}[htb]
\begin{center}
\includegraphics[width=\linewidth]{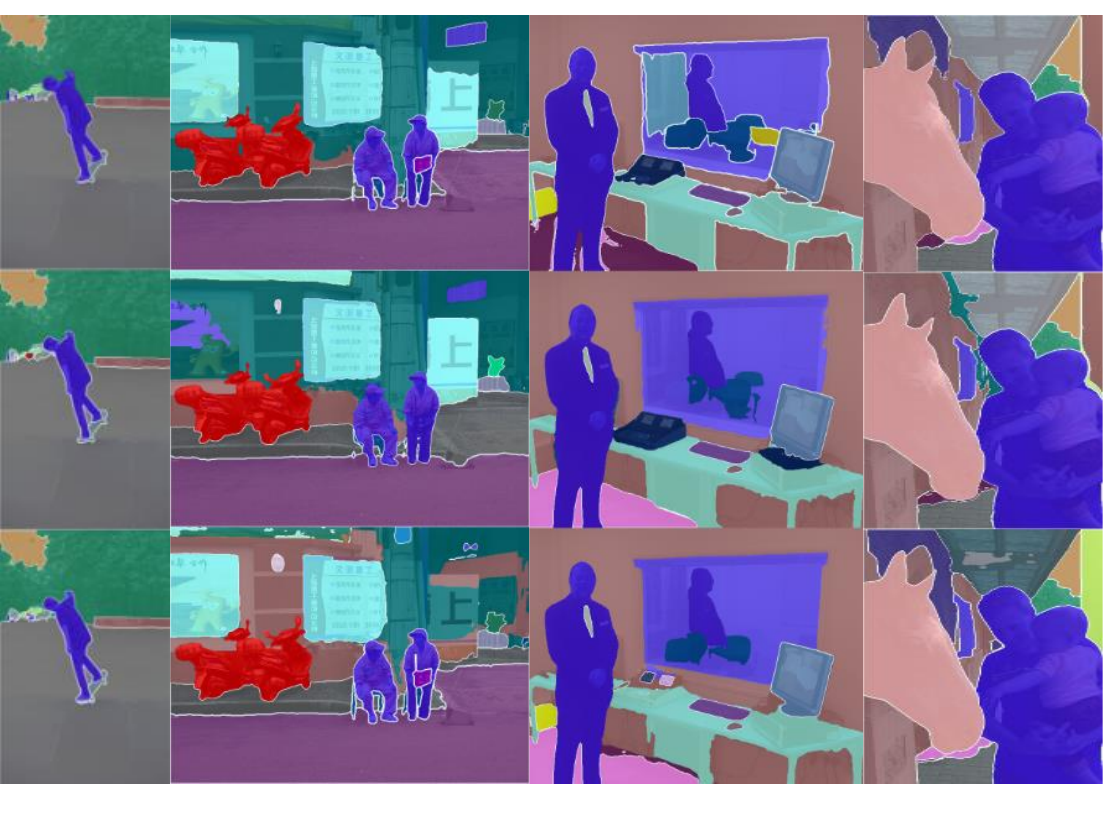}
\end{center}
\caption{\textbf{Comparison between U2Seg, STEGO, and S2-UniSeg on Unsupervised Semantic Segmentation.} (top) Visualizations of Semantic Segmentation of U2Seg. (middle) Visualizations of Semantic Segmentation of STEGO. (bottom) Visualizations of Panoptic Segmentation of S2-UniSeg.}
\label{fig:sem}
\end{figure*}


\section{D. Datasets details}
\label{sec:dataset}
\noindent\textbf{ImageNet and SA-1B.} Following previous works~\cite{cutler, u2seg}, we use ImageNet-1k to pretrain our model. ImageNet-1k has a total of 1.3M images. We also follow the recent work~\cite{unsam} to pretrain our model on subset of SA-1B~\cite{sam}. We use the subset of 0.4M, 1.2M, and 2.0M images from SA-1B to pretrained our model and compare with the recent UnSAM on class-agnostic instance segmentation.

\noindent\textbf{COCO.} The COCO dataset (Common Objects in Context)~\cite{coco} is extensively employed for universal segmentation research. It comprises 115,000 annotated training images, 5,000 annotated validation images, and over 200,000 unlabeled images. The dataset provides instance-level, semantic-level, and panoptic segmentation annotations. Following previous works, to assess our model's performance on class-agnostic instance segmentation, we conducted evaluations on the COCO Val2017 subset, consisting of 5,000 validation images, without utilizing any images from the COCO training set for model training or fine-tuning. 

\noindent\textbf{COCOStuff-27.} COCO-stuff is a large-scale dataset that consists of dense pixel-wise classification annotations. We follow STEGO~\cite{stego} and U2Seg~\cite{u2seg} to evaluate our model on semantic segmentation. COCOStuff-27~\cite{cocostuff} has 27 mid-level classes of the COCOStuff. We use unsupervised pixel accuracy and mean IoU as metrics.

\noindent\textbf{UVO.} The UVO dataset~\cite{uvo} is specifically curated for video object detection and instance segmentation tasks. Our experiments on unsupervised instance segmentation utilize the validation split of UVO, which comprises 256 videos annotated at a frame rate of 30 fps. Although our method does not require in-domain unsupervised COCO semantic segmentation as used by U2Seg, we follow them to exclude five additional non-COCO categories labeled as ``other'' in the official annotations for fair comparisons. Performance evaluation follows standard COCO-style metrics.

\noindent\textbf{Cityscapes.} The Cityscapes dataset~\cite{cityscape} is a well-established benchmark designed to advance semantic understanding of urban environments, with a primary focus on semantic segmentation of street scenes. For our study on unsupervised universal image segmentation, we employ the validation splits of Cityscapes. Evaluation follows COCO-style panoptic metrics, adhering to standard benchmarking practices.

\section{E. Training details}
\noindent\textbf{Model Architecture.} Following previous works~\cite{u2seg, cutler, unsam}, we use the DINO pretrained ViT-base/8 (hidden dimension of 768 and 12 self-attention layers) as our backbone. We augment the original DINO with light-weight ViT-Adapter~\cite{vit_adapter} to provide multi-scale information, where an Interaction block~\cite{vit_adapter} is inserted every 4 DINO blocks. Only Deformable Attention~\cite{mask2former} is used as self/cross attention for the Interaction block to achieve linear complexity. For the mask decoder, we use the official setting of Mask2Former~\cite{mask2former} of using 6 decoder layers, where each decoder is a stack of masked cross-attention, self-attention, and fully-connected layers. The original classification head is replaced by the projection head in our paper, where we follow DINO~\cite{dino} using a 3-layer MLP with hidden dimension 2048 followed by L2 normalization and a linear layer of $K$ dimensions. The teacher and student queries are both query feats before the LayerNorm layer.
After training, the final teacher encoder with the mask decoder is used for inference. We follow the official Mask2Former semantic/instance/panoptic inference method. For example, for semantic segmentation, the final semantic mask for each class is the Einstein summation of query feats, query class probabilities, query masks ('nc,ck-\>nk; nk,nhw-\>hwk').

\noindent\textbf{Feature Pyramid.} Since image segmentation often has multi-scale training~\cite{fpn} where a large size such as 768 is used, we resize the image as half before input to the DINO branch, as if we are using a patch size of 16. The Adapter branch takes the original image as input, so S2-UniSeg also has a pyramid of 4, 8, 16, 32 as previous works~\cite{u2seg, unsam}.

\noindent\textbf{UniAP is non-parametric.} It should be noted that our model, compared with U2Seg, does not require semantic segmentation pre-training. Specifically, U2Seg uses the pretrained STEGO~\cite{stego} model to generate the semantic-level segmentation masks. While our method for UniAP is non-parametric and does not require pretraining to generate semantic-level pseudo-masks.

\noindent\textbf{Hyper-parameter setting.} We set $\sigma=0.07, \{\tau_t\}_{t=1}^{\Gamma}=[0.8, 0.7. 0.6, 0.5, 0.4], \phi=5, \omega_f=0.6, \omega_s=0.4, K=512$. The number of semantic and instance queries are set to 50 and 150. A local multi-crop scale between 0.05 and 0.4 is used for multi-crop training. 

\noindent\textbf{Optimization setting.} We train the model using AdamW with a batch size of 16 on 8 A800 GPUs. The learning rate is linearly warmed up to 0.000625 for 10k iterations. Our model is trained for 160k iterations, while other CutLER models are additionally self-trained for several 80K iterations. We use a cosine schedule for the teacher momentum to change from 0.996 to 1.

\section{F. Preliminaries of Bipartite matching post-processing for class-aware universal segmentation.}
\label{sec:match}
We follow U2Seg~\cite{u2seg} to post-process the pseudo-classes ($K$ in our paper) using bipartite matching and evaluate our method on class-aware benchmarks same as U2Seg. Specifically, our model has $K$ pseudo-classes (also called clusters), which is the output dimension of the projection head. Suppose the evaluated benchmark has $C$ classes (such as 27 ``stuff'' classes in COCOStuff27 semantic segmentation, 80 ``thing'' classes in COCO instance segmentation, or merged ``thing'' and ``stuff'' classes in panoptic segmentation). During evaluation, we follow STEGO~\cite{stego} to use Hungarian matching to find the best subset of $C$ pseudo-classes out of the $K$ pseudo-classes as the class logits. We use softmax on these $C$ matched pseudo-classes to compute their class probabilities. 


\section{G. Complete quantitative results on unsupervised instance, panoptic, and class-agnostic instance segmentation}
\label{sec:full}

In this section, we provide the comprehensive evaluated results on each benchmark, including class-agnostic instance segmentation, instance segmentation, and panoptic segmentation. For these 3 benchmarks, we use official COCO API to provide the full results of standard COCO metrics, including instance-level AP across various IoU thresholds - AP (averaged over IoU thresholds from 0.5 to 0.95 with a step size of 0.05), AP50 (IoU@0.5) and AP75 (IoU@0.75), and AP across scales - APS (small objects), APM (medium objects) and APL (large objects), and panoptic-level ``thing"/"stuff'' panoptic quality $\text{PQ}$, $\text{PQ}^{\text{Th}}$, $\text{PQ}^{\text{St}}$, segmentation ``thing"/"stuff'' quality $\text{SQ}$, $\text{SQ}^{\text{Th}}$, $\text{SQ}^{\text{St}}$, and recognition ``thing"/"stuff'' quality $\text{RQ}$, $\text{RQ}^{\text{Th}}$, $\text{RQ}^{\text{St}}$. The full results on Unsupervised Panoptic Segmentation on COCO and Cityscapes are presented in Table~\ref{tab:full_pano}. The full results on Unsupervised Class-agnostic Instance Segmentation on COCO are presented in Table~\ref{tab:full_cls_ag}. The full results on Unsupervised Instance Segmentation on COCO and UVO are presented in Table~\ref{tab:full_ins}. 

Note that for unsupervised semantic segmentation on COCOStuff-27, previous works including STEGO~\cite{stego} also reported the ``Linear Probe'' mean IoU and pixel accuracy, in the sense that ground truth class labels are used to back-propagate the gradient to the \textit{detached} linear classifier inputs. Our method and U2Seg do not use linear probe and only report the unsupervised mean IoU and pixel accuracy.

\newpage
\begin{table*}[hbt]
\small
\begin{center} 
\begin{tabular}{l|l|ccccccccc}
 Dataset  & Pretrain Dataset  & PQ      & PQ$^\text{St}$     & PQ$^\text{Th}$     & SQ      & SQ$^\text{Th}$     & SQ$^\text{St}$    & RQ       & RQ$^\text{Th}$      & RQ$^\text{St}$     \\ \hline
 \multirow{4}{*}{COCO} &ImageNet       & 20.2 & 19.8  & 26.1 & 80.6 & 80.9 & 78.4 & 26.7   & 24.8 & 36.8 \\
       &SA-1B-0.4M                     & 25.7 & 24.6 & 30.4 & 88.4 & 89.1 & 86.7 & 28.6   & 27.0 & 39.4 \\
     &SA-1B-1.2M                       & 28.3 & 28.1 & 32.8 & 89.2 & 90.4 & 88.5 & 29.3   & 28.4 & 41.0 \\
     &SA-1B-2.0M                       & 29.6 & 28.4 & 34.2 & 90.3 & 90.7 & 89.4 & 30.2   & 29.8 & 41.8 \\ \hline 
\multirow{4}{*}{Cityscapes} &ImageNet  & 20.5 & 11.0  & 25.8 & 60.1 & 61.0 & 59.8 & 29.7   & 16.5  & 41.2 \\
       &SA-1B-0.4M                     & 22.5 & 12.4  & 27.4 & 63.7 & 64.1 & 61.9 & 32.6   & 18.4  & 42.8 \\
     &SA-1B-1.2M                       & 23.6 & 14.6  & 28.1 & 64.3 & 64.8 & 63.0 & 33.4   & 18.8  & 44.6 \\
     &SA-1B-2.0M                       & 25.4 & 15.4  & 29.7 & 66.7 & 67.9 & 64.8 & 35.0   & 20.8  & 47.2 \\ \hline  
\end{tabular}\vspace{-1.2em}

\end{center}
\caption{\textbf{Complete results for unsupervised panoptic segmentation on COCO\texttt{val2017} and Cityscapes\texttt{val}.} We show results of our S2-UniSeg pretrained on different dataset splits, including ImageNet, and SA-1B-0.4M, SA-1B-1.2M, and SA-1B-2.0M.}
\label{tab:full_pano}
\end{table*}

\begin{table*}[hbt]
\small
\begin{center} 
\begin{tabular}{l|lcccccccccc}
 Pretrain Dataset  & $\text{AP}^{\text{mask}}_{\text{50}}$      & $\text{AP}^{\text{mask}}_{\text{75}}$     & $\text{AP}^{\text{mask}}$     & $\text{AP}^{\text{mask}}_{\text{S}}$      & $\text{AP}^{\text{mask}}_{\text{M}}$     & $\text{AP}^{\text{mask}}_{L}$    & $\text{AR}^{\text{mask}}_{1}$    & $\text{AR}^{\text{mask}}_{10}$     & $\text{AR}^{\text{mask}}_{100}$  \\ \hline
 ImageNet   & 23.4 & 13.8  & 14.2 & 8.4 & 14.4 & 19.2 & 11.6   & 20.0 & 34.3 \\
 SA-1B-0.4M & 41.6 & 33.4 & 33.6 & 28.0 & 31.8 & 38.9 & 30.2   & 40.8 & 43.9 \\
 SA-1B-1.2M & 44.9 & 34.8 & 35.0 & 30.2 & 34.8 & 40.5 & 32.5   & 42.7 & 44.6 \\
 SA-1B-2.0M & 48.7 & 35.6 & 36.4 & 30.8 & 35.9 & 42.1 & 34.8   & 43.6 & 46.0 \\ \hline  
\end{tabular}\vspace{-1.2em}

\end{center}
\caption{\textbf{Complete results for unsupervised class-agnostic instance segmentation on COCO\texttt{val2017}.} We show results of our S2-UniSeg pretrained on different dataset splits, including ImageNet, and SA-1B-0.4M, SA-1B-1.2M, and SA-1B-2.0M.}
\label{tab:full_cls_ag}
\end{table*}

\begin{table*}[hbt]
\small
\begin{center} 
\begin{tabular}{l|c|ccccccccc}
 Dataset  &Pretrain Dataset& $\text{AP}^{\text{mask}}_{\text{50}}$      & $\text{AP}^{\text{mask}}_{\text{75}}$     & $\text{AP}^{\text{mask}}$     & $\text{AP}^{\text{mask}}_{\text{S}}$      & $\text{AP}^{\text{mask}}_{\text{M}}$     & $\text{AP}^{\text{mask}}_{L}$    & $\text{AR}^{\text{mask}}_{1}$    & $\text{AR}^{\text{mask}}_{10}$     & $\text{AR}^{\text{mask}}_{100}$  \\ \hline
 \multirow{4}{*}{COCO} &ImageNet       & 24.2 & 14.7  & 15.3 & 8.7 & 14.9 & 16.8 & 9.3   & 20.4 & 35.5 \\
       &SA-1B-0.4M                     & 34.0 & 26.4 & 27.4 & 19.4 & 26.8 & 30.0 & 19.8   & 32.7 & 36.1 \\
     &SA-1B-1.2M                       & 35.5 & 28.4 & 29.0 & 22.5 & 28.1 & 31.4 & 21.4   & 34.7 & 38.4 \\
     &SA-1B-2.0M                       & 38.4 & 29.1 & 30.7 & 24.6 & 30.4 & 32.5 & 24.7   & 36.1 & 40.3 \\ \hline 
\multirow{4}{*}{UVO} &ImageNet         & 27.3 & 15.4  & 16.2 & 9.2 & 15.1 & 19.4 & 9.7   & 24.5  & 32.0 \\
       &SA-1B-0.4M                     & 31.2 & 19.8  & 20.8 & 14.5 & 18.7 & 25.9 & 15.5 & 30.0  & 34.6 \\
     &SA-1B-1.2M                       & 32.4 & 22.4  & 22.6 & 16.4 & 20.8 & 27.4 & 15.8   & 31.7  & 35.7 \\
     &SA-1B-2.0M                       & 33.0 & 23.4  & 24.3 & 17.9 & 22.6 & 29.4 & 17.1   & 32.5  & 36.4 \\ \hline
\end{tabular}
\end{center}
\caption{\textbf{Complete results for unsupervised instance segmentation on COCO\texttt{val2017} and UVO\texttt{val}.} We show results of our S2-UniSeg pretrained on different dataset splits, including ImageNet, and SA-1B-0.4M, SA-1B-1.2M, and SA-1B-2.0M.}
\label{tab:full_ins}
\end{table*}

\end{document}